\documentclass{article}
\usepackage{spconf,amsmath,graphicx}
\usepackage[nolist,nohyperlinks]{acronym}
\usepackage{multirow}
\usepackage[pdfencoding=auto]{hyperref}

\title{{\it\bf Oh, Jeez!} or {\it \bf uh-huh}? \\ A listener-aware Backchannel predictor on ASR transcriptions}
\name{Daniel Ortega, Chia-Yu Li, Ngoc Thang Vu 
}
\address{Institute for Natural Language Processing (IMS)\\ University of Stuttgart, Germany \\
            \{daniel.ortega, chia-yu.li, thang.vu\}@ims.uni-stuttgart.de}

\begin{document}
%
\maketitle
\begin{abstract}
This paper presents our latest investigation on modeling backchannel in conversations. Motivated by a proactive backchanneling theory, we aim at developing a system which acts as a proactive listener by inserting backchannels, such as continuers and assessment, to influence speakers. Our model takes into account not only lexical and acoustic cues, but also introduces the simple and novel idea of using listener embeddings to mimic different backchanneling behaviours. Our experimental results on the Switchboard benchmark dataset reveal that acoustic cues are more important than lexical cues in this task and their combination with listener embeddings works best on both, manual transcriptions and automatically generated transcriptions.
\end{abstract}

\section{Introduction}
\label{sec:intro}

In multi-party conversations, interlocutors interact in such a way, that one of them speaks (speaker) and the rest (listeners) listen, interchanging the speaker-listener roles throughout the dialog. However, the interaction is actually more complex including \ac{bc} responses\footnote{The terms backchannel and backchannel responses are used indistinctly.} from the listeners, i.e. signals of acknowledgment like {\it uh-huh},  or a particular reaction, e.g. {\it oh, jeez!}, to what the speaker just uttered  \cite{Bangerter2003}.

The term backchannel was coined by Yngve \cite{Yngve1970}, who divides the dialog into frontchannel and backchannel; the former corresponds to the channel of the interlocutor who currently has the floor, and the latter is the channel that does not. \ac{bc} responses, made by the listeners, do not occur in separate turns, but during the speaker's turn \cite{Bangerter2003}. Non-verbal \acp{bc} also exist like nodding, gestures, smiling \cite{brunner1979, bavelas2011, bertrand2007}, however, this type of \acp{bc} are out of our research scope.

Different theories have been developed to explain human-human dialog interaction and they approach \ac{bc} responses differently. Tolins and Fox \cite{tolins2014} cluster backchanneling studies into two theory paradigms:

\noindent{\bf Reactive backchanneling theory:}  Under this view,  speech comprehension and production are treated as independent processes. Listeners are considered as passive recipients of information, using \acp{bc} only to display the acceptance of speakers’ planned multi-turn conversation, so that \acp{bc} are used as supportive messages, but they do not play a central role \cite{clancy1996}.

\noindent{\bf Proactive backchanneling theory:} Contrary to the previous theory, speech comprehension and production complement each other. Listeners are active in the construction of the dialog by using \acp{bc}. Consequently, the speaker should not only take care of their talk, but also monitor listeners' reactions, i.e. \ac{bc} responses, and adjust their talk accordingly in order to accomplish a successful conversation \cite{clark2004}. Our research is aligned with this theory. 

Several works \cite{tolins2014, schegloff1982, goodwin1986} have explored the type of information that \acp{bc} yield. Theoretical research on \ac{bc} has branched into two main paths. On the one side, it has focused on explaining \ac{bc} functionalities and categorizing them. On the other side, it has focused on \ac{bc} placement, i.e. where within the speaker's speech the \acp{bc} occur (See \cite{tolins2014} for further details). 

Our research takes the functional approach and the categorical distinction between generic and specific \acp{bc} \cite{goodwin1986}, defined as follows:

\noindent{\bf Generic \acp{bc}}, also called continuers \cite{schegloff1982}, exhibit acknowledgement and understanding of what the speaker said, as well as encourage speaker to continue. Examples are {\it uh-huh} and {\it mm-hm}. 

\noindent{\bf Specific \acp{bc}}, also called assessment \cite{goodwin1986}, are contextualized responses, so that the listener reacts to what the speaker uttered without intending to take the floor. Examples are {\it oh, jeez!} and {\it yeah}.

From the computational point of view, human conversational features, like backchanneling, should be explored and integrated in voice-driven appliances, such as virtual assistants and smart speakers to create a more human-like experience. 
Up to date, research on computational modeling of backchanneling \cite{ward1996using, ward2000prosodic, morency2010probabilistic, huang2010learning, ruede2017} has only considered backchanneling as a binary decision ({\it yes} or {\it no}). 
In this work, we present a neural-based model that takes lexical and acoustic features as input for \ac{bc} prediction from three categories: No-\ac{bc}, \ac{bc}-Continuer and \ac{bc}-Assessment aligning with the aforementioned theory. To the best of our knowledge, we are the first to extend the modeling task of backchanneling to the proactive level with more fine-grained categories of backchannels. Although in the literature, it is proven that backchanneling behaviour is only language dependent \cite{heinz2003}, we believe that this behaviour is also speaker dependent within a language. Therefore, we propose a simple efficient method to encode listener behaviour as an embedding in our model which led to improvements of the \ac{bc} predictor's performance. 

Furthermore, all work on automatic \ac{bc} prediction has been done using only \ac{mt}.
Nonetheless, this type of data differs substantially from real data, i.e. \ac{at}, generated by \ac{asr} systems.
Therefore, in this paper we explore the effect of training and testing the proposed model on \ac{at}. Our goal at this point is to take \ac{bc} prediction into a more realistic scenario.

In sum, we introduce a \ac{bc} predictor that process both lexical and acoustic information from the frontchannel and also encodes the listener behaviour. 
We train and test our model on different scenarios to contrast the effect of using manual and automatically generated transcriptions from a hybrid \ac{tdnn}/\ac{hmm} \ac{asr} system. 
Our results show that a) both lexical and acoustic cues are crucial for this task but acoustic cues seem to play a more important role and b) the listener embedding improves consistently the accuracy of the model performance on the benchmark dataset \ac{swbd}\cite{godfrey1992}. 

\section{Backchannel prediction}
\label{sec:bc_prediction}
\begin{figure*}
    \centering
    \includegraphics[width=0.75\textwidth]{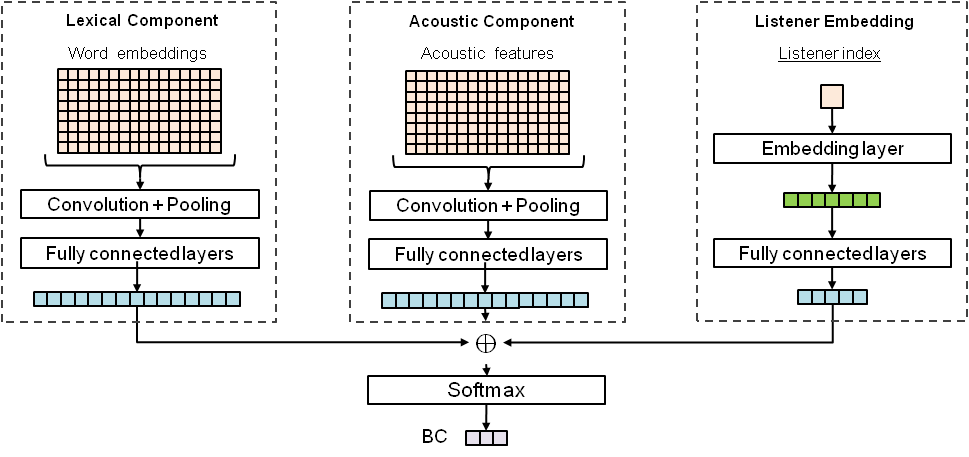}
    \caption{Backchannel predictor's model. $\oplus$ represents a concatenation.}
    \label{fig:model}
\end{figure*}

Our proposed model for \ac{bc} prediction, depicted in \autoref{fig:model} consists of three parts that work in parallel: a \ac{cnn} that generates vector representation from the lexical information, another a \ac{cnn} that generates vector representation from the acoustic features, and an embedding layer followed by a \ac{ffn} that encodes the speaker behaviour.

The lexico-acoustic component, i.e. the bi-\ac{cnn} constituent, is based on \cite{ortega2018}.  
The lexical input consists of a grid-like representation made by stacking up the word embeddings of the $n$ previous speaker's words given a certain time point in the conversation. 
Similarly, given that time point, acoustic features at frame level of the last 2000 ms are extracted from the frontchannel and stacked up. 
Each input is processed by a \ac{cnn} that generates a vector representation.

\acp{cnn} perform a discrete convolution using a set of different filters on an input matrix, where each column of the matrix is the word embedding of the corresponding word.
We use 2D filters $f$ (with width $|f|$) spanning over all embedding dimensions $d$ as described by the following equation:

\begin{equation}
	(w \ast f)(x,y) = \sum_{i=1}^{d}\sum_{j = -|f|/2}^{|f|/2}w(i,j) \cdot f(x-i,y-j)
\end{equation}

The third model's input is single integer, that represents the unique identifier for the listener. That integer is passed to a embedding layer, that stores the listeners' vector representation, followed by two \acp{ffn}. Note that lexical and acoustic information are not aligned as in other works. Our intuition is that the lexical meaning and the acoustic cues should be processed separately, so that both sources are not constrained during the training.
Finally, the three resulting vectors are concatenated and passed to a softmax layer, that outputs a probability distribution over the three classes: No-\ac{bc}, \ac{bc}-Continuer and \ac{bc}-Assessment.





\section{Experimental setup}
\label{sec:experimental_setup}

\subsection{Data and setup for \ac{bc} prediction}
\label{ssec:bc_setup}

\subsubsection{Backchannel data annotation}
\label{sssec:lex_annotation}
We evaluate our model on \ac{swbd} \cite{godfrey1992}, a dialog corpus of telephone conversations. 
Annotations and time stamps for \ac{bc} are taken from  \cite{jurafsky1997switchboard}, and further expanded by following the setup proposed by Ruede et al. \cite{ruede2017}. 
We also took the time stamps for No-\ac{bc} instances from \cite{ruede2017} as well as the spĺits. 
From a total of 2438 conversations, 2000 are for training, 200 for validation and 238 for test.
As result, our data consists of almost 61k \ac{bc} instances and the same amount for No-\ac{bc}. 

However, we decided to take the classification to be consistent with the \ac{bc} categorization proposed in \cite{goodwin1986}, i.e continuer vs assessment \acp{bc}. Since the \ac{swda} data is not annotated in such a way, we manually annotated them with continuer vs assessment \acp{bc} labels.

The annotation process was as follows, we extracted a list of 670 unique utterances used within the corpus to backchannel. Then, according to the theory, continuers \cite{goodwin1986, Bangerter2003, tolins2014}, aka general \acp{bc}, are {\it uh-huh}, {\it um-hum} and their variations. 
Following that, we annotated as continuers those instances, whose realization contained exclusively these words or their variations, while ignoring filter noise or other markers; the rest was annotated as assessment. 
As result, only 68 out of 670 unique \ac{bc} realizations were annotated as continuers. 
However, the distribution in the dataset among both \ac{bc} classes is almost balanced: No-\ac{bc}  50\%, \ac{bc}-Continuer 22.5\% and \ac{bc}-Assessment 27.5\%. 

For listener annotation, we obtain from \ac{swbd} corpus the mapping between dialog channels and unique speakers. On average, each speaker takes part in 10 conversations. Although the official \ac{swbd} documentation reports 543 speakers, the annotation did not include speakers for 11 channels. Pragmatically, we linked those channels to a random speaker. In total, our model learns 519 listener embeddings.

\subsubsection{Backchannel data extraction}
\label{sssec:ac_data}

As mentioned in \autoref{sssec:lex_annotation}, we obtained the time stamps for all instances in the dataset, that were later used to extract both lexical and acoustic features. 

Lexical information refers to 15 previous words from the frontchannel that happened before the time stamp that marks the instance in the backchannel. We used setup from \cite{ruede2017} to extract the words from \ac{mt}. \ac{at} were generated by our \ac{asr} system described in \autoref{ssec:asr_setup}.

Time stamps are also used to obtain acoustic features of the 2000 ms from the frontchannel speech signal before the \ac{bc} happens. Acoustic features are extracted at frame level, i.e. the speech signal is divided into frames of 25 ms with a shift of 10 ms. Two types of acoustic features were extracted using the openSMILE toolkit \cite{tools:openSMILE}: 3 prosodic features (fundamental frequency, loudness and voicing probability), 13 \acp{mfcc}. 


\subsection{Data and setup for \ac{asr}}
\label{ssec:asr_setup}

We employed KALDI \cite{Kaldi} to build the hybrid \ac{tdnn}/\ac{hmm} ASR system \cite{waibel1989phoneme, Povey2016PurelySN}. 
To the extent of our knowledge, it is one of the best hybrid \ac{asr} systems available for research and thus was selected for our experiments.
In the recipe, 40 \acp{mfcc} were computed at each time step and each frame was appended to a 100-dimensional iVector. Speaker adaptive feature transform techniques and data augmentation techniques were implemented. The alignments are based on the \ac{gmm}/HMM model with linear discriminant analysis (LDA) \cite{lda}, maximum likelihood linear transform (MLLT) \cite{mllt} and speaker adaptive training (SAT) \cite{sat1,sat2}.  Besides, the language mode is 3-gram  trained with the transcriptions. The word error rate (WER) of this hybrid \ac{tdnn}/\ac{hmm} system on train, validation and test sets are 16.88\%, 16.89\% and 20.51\%, respectively. 


		

\section{Experimental results}
\label{sec:experimental_results}
\begin{table}[ht]
\footnotesize{}
\centering
\begin{tabular}{|l|c|c|}
\hline 
\textbf{Hyperparameter}		&\textbf{LM} &\textbf{AM}\\ 
\hline
    Filter widths 				&3, 4, 5 & 11, 12\\
    Number of filters		&16 &16 \\
    Dropout rate				&0.5 &0.5 \\
    Activation function 		&ReLU&ReLU  \\
    Pooling size		 		&(50, 1)&(18, 1)\\
    Word embeddings				&word2vec \cite{WordEmbeddings:word2vec}&--- \\
    Acoustic features				& --- &Prosodic \& MFCC \\
    Number of frames				& --- &148 \& 198 \\
    Mini-batch size				&64 &64  \\
\hline
\end{tabular}
\caption{\label{tab:hyperparams} Hyperparameters. }
\end{table}

We present the results from three models, i.e. the acoustic model that takes only acoustic features, the lexical model that only takes word embeddings and the lexico-acoustic model (\autoref{fig:model}) that takes both inputs and the listener embedding. Their hyperparameters are presented in Table\ref{tab:hyperparams}. The listener embedding is a 5-dimension vector. 

\subsection{Acoustic model}

\autoref{tab:ac_results} summarizes the results obtained from training and testing only the acoustic component on prosodic and MFCC features at with different number of frames, i.e. 148 and 198 frames representing 1.5 s and 2 s, respectively. In both scenarios, MFCC features yield better results. The best performance occurs when 148 frames are fed into the acoustic model. Therefore, we experimented further only using MFCC features and 148 frames. 

\begin{table}[h!]
\footnotesize{}
\centering
		\begin{tabular}{|l|c|c|}
			\hline 
			\multicolumn{1}{|c|}{\multirow{2}{*}{\textbf{Time}}}& \multicolumn{2}{c|}{ \textbf{Features}}\\
			\multicolumn{1}{|c|}{}	&\bf Prosodic	&\bf MFCC	\\
            \hline
			1.5 s (148 frames)	&53.7	&\bf 54.7	\\
			2.0 s (198 frames)	&52.9	&54.2	\\
			\hline
		\end{tabular}
	\caption{\label{tab:ac_results} Accuracy results (\%) on acoustic model.}
\end{table}

\subsection{Lexical model}

Experiments with the lexical component had the goal of finding the appropriate number of words from the frontchannel. We tested from 5-10 and the best performance was found with 5 words, in both scenarios - \ac{mt} and \ac{at} (see \autoref{tab:lex_results}). This hyperparameter was then used for further experiments. 

\begin{table}[h!]
\footnotesize{}
\centering
		\begin{tabular}{|c|c|c|}
			\hline 
			\multicolumn{1}{|c|}{\multirow{2}{*}{\textbf{Words}}}& \multicolumn{2}{c|}{ \textbf{Transcriptions}}\\
			\multicolumn{1}{|c|}{}	&\bf Manual	&\bf Automatic	\\
            \hline
			5	&\bf 53.9	&\bf53.7	\\
			6	&53.6	&53.1	\\
			7	&52.2	&53.0	\\
			\hline
		\end{tabular}
	\caption{\label{tab:lex_results} Accuracy results (\%) on lexical model.}
\end{table}

\subsection{Lexico-acoustic model}
\label{ssec:lex_ac_res}

\autoref{tab:all_results} summarizes our experimental results in accuracy terms obtained from three models (acoustic, lexical and lexico-acoustic) in different configurations: no listener vs. with listener and \ac{mt} vs. \ac{at}. Models were tested on the same type of transcriptions used for training, no cross-transcriptions was performed. However, experimental results in \cite{ortega2018} suggest to train and test on the same type of data.

\begin{table}[h!]
\footnotesize{}
\centering
		\begin{tabular}{|l|c|c|c|c|}
			\hline 
			\multicolumn{1}{|c|}{\multirow{2}{*}{\textbf{Model}}}& \multicolumn{2}{c|}{ \textbf{No listener}}& \multicolumn{2}{c|}{ \textbf{ With listener}}\\
			\multicolumn{1}{|c|}{}	&\bf\ac{mt}	&\bf\ac{at}	&\bf \ac{mt}	&\bf\ac{at}\\
            \hline
			Acoustic	&\multicolumn{2}{c|}{54.7}	&\multicolumn{2}{c|}{57.1}\\
			Lexical	        &53.9	&53.7	&56.3	&55.8\\
			Lexico-acoustic	&56.7	&56.4	&\bf 58.9	&\bf58.6\\
			\hline
		\end{tabular}
	\caption{\label{tab:all_results} Global accuracy results (\%).}
\end{table}

\section{Discussion \& Analysis}
\subsection{Discussion}
Overall, our results indicate that the combination of lexical and acoustic information leads to the best performance, reaching 58.9\% when the model was trained  on \ac{mt} and the listener embedding was present. Furthermore, listener embeddings contribute to a consistent improvement. 

A surprising finding is that the acoustic model outperforms the lexical one, regardless of the use of \ac{mt}	or \ac{at} in the latter model. This contradicts the findings from dialog act tagging \cite{ortega2018}, where lexical information carries more important cues, however, both tasks are not completely comparable.

Another interesting finding is that the performance difference in the lexical model between \ac{mt}	or \ac{at} is relatively small, although the WER in the ASR system ranges from 16.9\% to 20.5\%. A deeper analysis is still needed to find out which lexical information is valuable in the ASR transcriptions.

\subsection{Analysis}
\begin{figure}
    \centering
    \includegraphics[width=0.4\textwidth]{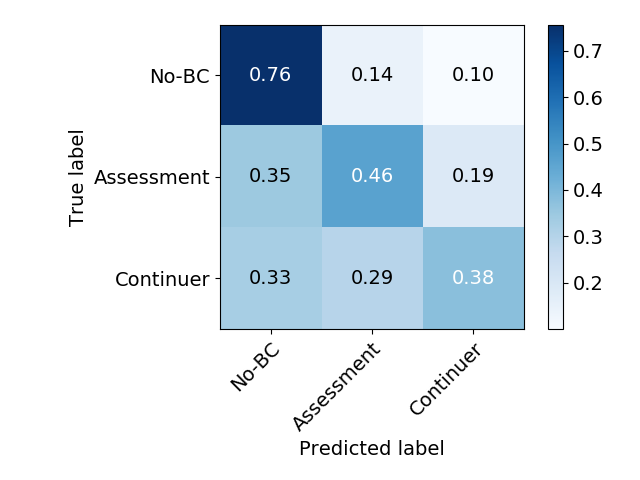}
    \caption{Confusion matrix for the best results on ATs}
    \label{fig:CM_norm}
\end{figure}

In order to understand deeper our results and their implications, we present a confusion matrix (\autoref{fig:CM_norm}) of the lexico-acoustic with listener embedding on \ac{at}.


Our model failed at predicting No-BC almost the half of the time, instead of the two other classes. We can assume that the model, seen as a listener, was acting tentative to backchannel, so that one of the future work could be to introduce a parameter in the model to control the backchanneling rate.

Focusing on \ac{bc} classes, the model was able to distinguish between assessment and continuer. However, our system failed more at predicting continuers. A deeper research at lexical and prosodic levels should be carried out, in order to understand the contextual differences that trigger each of them.

\section{Conclusion}
We investigated modeling backchanneling from listener perspectives in conversations motivated by proactive backchanneling theory, which suggests different categories of backchanneling and how they influence speakers. We proposed a model that combines lexical and acoustic cues and introduces a simple but novel method of using listener embeddings to mimic different backchanneling behaviours. Our experimental results on the Switchboard benchmark data set show several interesting findings: 1) acoustic cues are more important than lexical cues in this task, 2) listener embeddings are useful to improve the overall performance and 3) our proposed model is robust against ASR errors.

\clearpage

\bibliographystyle{IEEEbib}
\bibliography{refs}

\begin{acronym}[Bash]
    \acro{adagrad}[AdaGrad]{adaptive gradient algorithm}
    \acro{am}[AM]{attention mechanism}
    \acro{crf}[CRF]{conditional random field}
    \acro{cnn}[CNN]{convolutional neural network}
    \acro{ffn}[FFN]{feed-forward network}
    \acro{da}[DA]{dialog act}
    \acro{dl}[DL]{deep learning}
    \acro{e2e}[E2E]{End-to-End}
    \acro{gd}[SGD]{stochastic gradient descent}
    \acro{hmm}[HMM]{hidden Markov model}
    \acro{nlp}[NLP]{natural language processing}
    \acro{rnn}[RNN]{recurrent neural network}
    \acro{svm}[SVM]{support vector machine}
    \acro{swbd}[SWBD]{Switchboard Corpus}
    \acro{swda}[SwDA]{The Switchboard Dialog Act Corpus}
    \acro{asr}[ASR]{automatic speech recognition}
    \acro{nn}[NN]{neural network}
    \acro{mt}[MTs]{manual transcriptions}
    \acro{at}[ATs]{automatic transcriptions}
    \acro{e2e}[E2E]{End-to-End}
    \acro{cd}[CD]{context-dependent}
    \acro{tdnn}[TDNN]{time-delay neural network} 
    \acro{ctc}[CTC]{connectionist temporal classification}
    \acro{wer}[WER]{word error rate}
    \acro{mfcc}[MFCC]{Mel-frequency cepstral coefficient}
    \acro{gmm}[GMM]{Gaussian Mixture Model}
    \acro{espnet}[ESPnet]{End-to-End Speech Processing Toolkit}
    \acro{bc}[BC]{backchannel}
\end{acronym}

\end{document}